\documentclass[lettersize,journal,twoside]{IEEEtran}

\IEEEoverridecommandlockouts

\newcommand{\ourhand}{{PLATO Hand}}

\title{\LARGE \bf
\ourhand{}: Shaping Contact Behavior with Fingernails \\for Precise Manipulation
}

\author{
        Dong Ho Kang$^{1}$,
        Aaron Kim$^{1}$,
        Mingyo Seo$^{1}$,
        Kazuto Yokoyama$^{2}$,
        Tetsuya Narita$^{2}$,
        and Luis Sentis$^{1}$

\thanks{
Manuscript received January 26, 2026; revised April 9, 2026; accepted May 4, 2026.
This paper was recommended for publication by Editor Júlia Borràs Sol upon evaluation of the Associate Editor and Reviewers' comments.
}

\thanks{$^{1}$ D.H. Kang, A. Kim, M. Seo, and L. Sentis are with The University of Texas at Austin, Austin, TX, USA (email: {\tt\small dongho@utexas.edu}).}%
\thanks{$^{2}$ K. Yokoyama and T. Narita are with Sony Group Corporation, Tokyo, Japan.}%

}

\usepackage{xcolor}
\usepackage{amsfonts}
\usepackage{amsmath}
\usepackage{amssymb}
\usepackage{graphicx}
\usepackage{tabularx}
\usepackage{booktabs}
\usepackage{dsfont}
\usepackage{enumitem}
\usepackage[T1]{fontenc}
\usepackage[utf8]{inputenc}
\usepackage{newtxtext,newtxmath}

\usepackage{cite}

\usepackage{algpseudocode}
\usepackage{algorithm}

\usepackage[hypertexnames=false]{hyperref}

\setlist[itemize]{leftmargin=*}

\newif\ifshowrevisions
\showrevisionsfalse

\ifshowrevisions
  \newcommand{\rev}[1]{\textcolor{black}{#1}}
  \newcommand{\rrev}[1]{\textcolor{blue}{#1}}
  \newcommand{\delrev}[1]{\textcolor{red}{#1}}
\else
  \newcommand{\rev}[1]{#1}
  \newcommand{\rrev}[1]{#1}
  \newcommand{\delrev}[1]{}
\fi

\begin{document}

\maketitle

\begin{abstract}
We present the PLATO Hand, a dexterous robotic hand with a hybrid fingertip that combines a rigid fingernail, embedded distal phalanx, and compliant pulp to shape contact behavior during manipulation. \rrev{By mechanically organizing how contact is initiated, supported, and transmitted at the fingertip, this structure creates stable and task-relevant contact conditions across diverse object geometries and grasp orientations.} We develop a strain-energy-based bending--indentation model to guide the fingertip design and to explain how material stiffness and contact geometry govern deformation partitioning within the fingertip.  \rrev{Experiments show improved pinch stability, improved fingernail-mediated dorsal-contact force transmission and proprioceptive observability}, and successful execution of edge-sensitive manipulation tasks, including paper singulation, card picking, and orange peeling. These results show that coupling a mechanically structured contact interface with a force-motion-transparent finger mechanism provides a principled approach to precise manipulation.
Our project page is at: {\tt\small\textbf{\url{https://platohand.github.io}}}.
\end{abstract}


\section{Introduction}
\label{sec:intro}

\IEEEPARstart{H}{uman-like} dexterity in robotic hands requires both kinematic versatility and effective force regulation at the contact interface. Early work focused on achieving diverse grasp postures and expanded reachable workspace within constrained form factors~\cite{utahmit,robonaut,gifuhand,dlrhand,acthand}. 
To accomplish this, designers employed complex mechanisms including tendon-driven transmissions~\cite{Kim2019FluidCapability,rollingcontacthand,orca,trx,pisa/iit,shadowrobotShadowDexterous}, spatial linkages~\cite{Kim2021IntegratedHand,linkagehand}, and serially stacked servo motors~\cite{dmanus,allegrohand,tesolloDG5FHumanoid}. 
While these approaches expanded kinematic capability, high transmission impedance and nonlinear force paths often make compliant, force-regulated interaction difficult to achieve~\cite{Wensing2017ProprioceptiveRobots, Sim2021TheRobots,Jeong2024BaRiFlex:Learning, romero2024eyesight, Lin2022AManipulations}. 
\rev{As a result, dexterous performance is often limited not only by how the hand moves, but by how contact is mechanically formed and regulated at the fingertip.}

Effective dexterous manipulation depends critically on how contact forces are established and modulated at the fingertip itself~\cite{okamuradext,rdexmani}. 
In human fingertips, this behavior is not governed by soft tissue alone, but by the coupled structure of the nail, nail bed, and underlying pulp, which constrains deformation and redirects force transmission pathways~\cite{Piraccini2014NailClinician, Kumagai2022ComparisonNails}. 
\rev{Rather than acting only as a hard distal edge, the nail provides dorsal reinforcement that suppresses undesired global deformation while preserving compliant local contact at the pulp.} Human studies further suggest that fingernail morphology measurably affects manipulation dexterity, with approximately 2 mm fingernail length reported to improve hand dexterity relative to 0 mm~\cite{Shirato2017EffectDexterity}.

Robotic fingernails have previously been used to assist thin-object pickup, edge engagement, or low-clearance grasping~\cite{fang2025dexop, Do2024DenseTact-Mini:Surfaces, jain2020nail, Odhner2013ACU, Murakami2003NovelManipulation}. 
However, these designs are often regarded primarily as grasp aids or rigid contact tips, rather than as structural features that influence fingertip stiffness distribution, deformation, and force transmission across a range of contact conditions. 
\rev{In this view, the role of the nail is not limited to assisting thin-object pickup, but extends to shaping contact behavior for stable, force-regulated manipulation.}

\begin{figure}[t]
  \centering
  \includegraphics[width=1\linewidth]{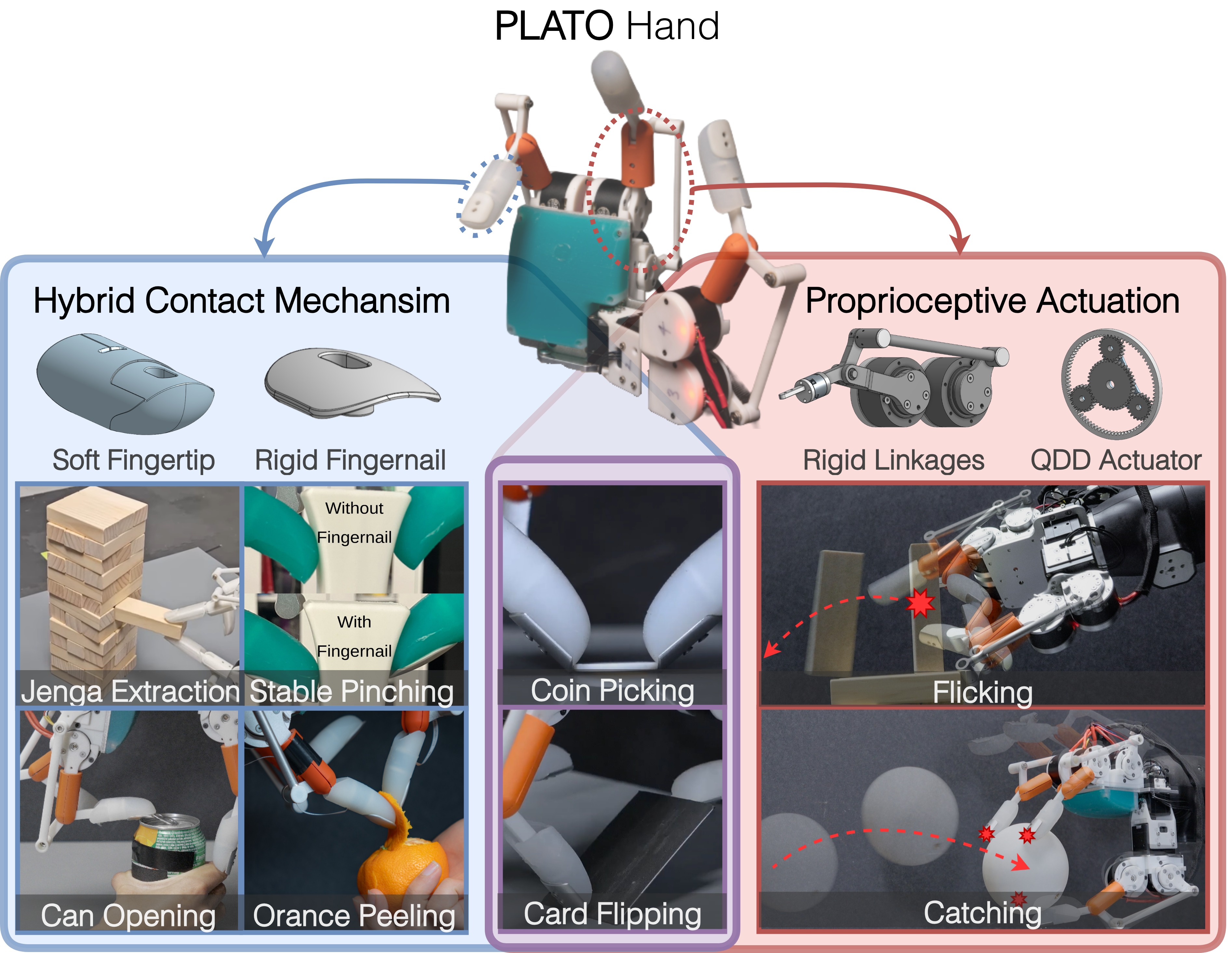}
  \caption{\textbf{Overview of the PLATO Hand}. 
  This system combines a hybrid fingertip with a rigid fingernail, distal phalanx, and compliant fingerpulp to shape how contact is formed and transmitted at the fingertip, together with proprioceptive actuation for high-bandwidth force-regulated interaction. By coupling localized edge interaction with force-motion transparency, the system enables robust and responsive behaviors across a range of precise dexterous manipulation tasks.}
  \label{fig:fig1}
\end{figure}

\begin{figure*}[t]
  \centering
  \includegraphics[width=1\linewidth]{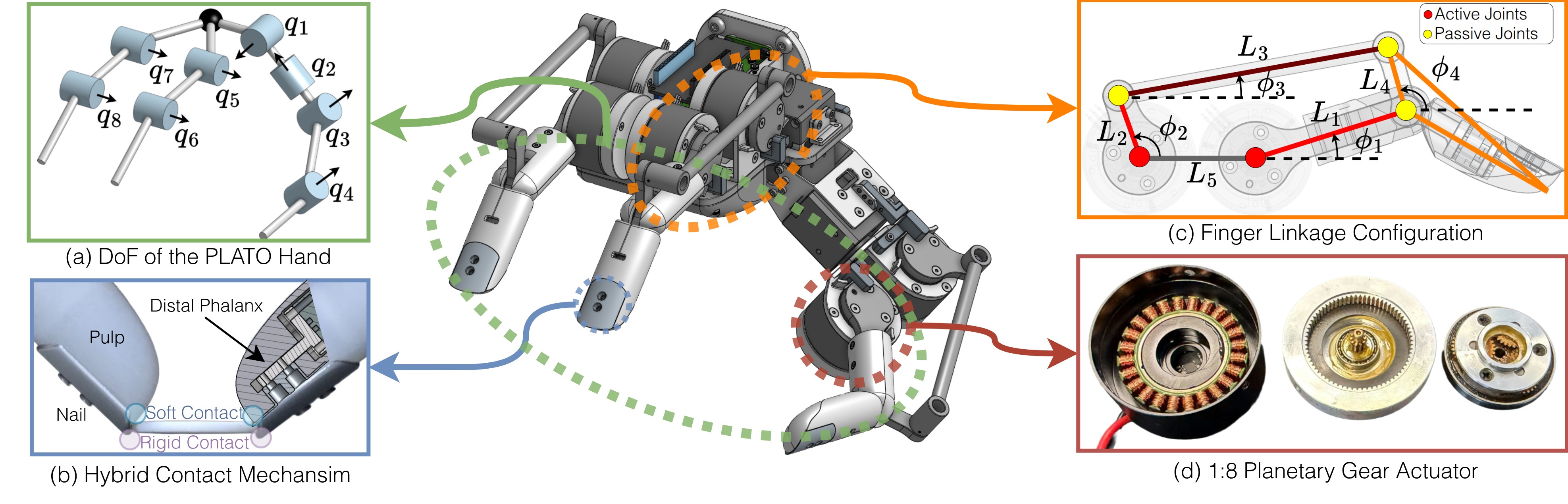}
  \caption{\textbf{Design overview of the PLATO Hand.}
  (a) Robot kinematic diagram of the hand with eight fully actuated joints across three fingers: two-DoF index and middle fingers, and a four-DoF thumb.
  (b) Cross-sectional view of the hybrid fingertip showing a rigid fingernail integrated with a compliant pulp surrounding the distal phalanx and a distal force–torque sensor. 
  (c) Five-bar linkage mechanism coupling the QDD actuators to the finger joints, enabling low-inertia transmission and configuration-dependent torque amplification.  
  (d) Internal structure of the 1:8 ratio planetary gearbox and Outrunner BLDC motor.
  }
  \label{fig:design}
\end{figure*}

To address these limitations, we present the PLATO (\mbox{\underline{P}}roprioceptive \mbox{\underline{L}}inkage-driven \mbox{\underline{A}}n\mbox{\underline{t}}hr\mbox{\underline{o}}pomorphic) Hand, which combines a low-impedance, proprioceptive finger mechanism with a hybrid fingertip \rev{composed of dorsal nail reinforcement, an embedded distal phalanx, and a compliant pulp. 
This hybrid structure mechanically organizes contact by shaping how stiffness, deformation, and force transmission are distributed within the fingertip.} 
By increasing effective flexural rigidity while providing internal support beneath the pulp, it suppresses undesirable global bending and preserves compliant local deformation at the contact interface, enabling both localized edge interaction and stable contact formation.
When combined with proprioceptive actuation using quasi-direct-drive (QDD) actuators~\cite{Wensing2017ProprioceptiveRobots}, these structural advantages become especially useful for high-fidelity force interaction in contact-rich manipulation.

In this paper, we introduce an energy-based model that captures deformation partitioning between global bending and local contact indentation in a hybrid fingertip architecture. 
The model reveals how material stiffness and contact curvature govern this partitioning and thereby shape contact behavior and stability. 
Based on this analysis, we design a fingertip and five-bar transmission mechanism that together support consistent force transmission across the workspace and reliable proprioceptive force sensing.
\rev{We further perform ablation experiments to isolate how the distal phalanx structure, fingernail, and compliant pulp each contribute to contact behavior across different object curvatures and contact configurations. 
These experiments reveal how each structural element contributes to deformation partitioning, contact stability, and the transmission and observability of contact-induced force variations.
}
The core contributions of this work are:
\begin{enumerate}[label=\roman*)]
\item The PLATO Hand, a proprioceptive robotic hand that integrates a hybrid fingertip comprising a fingernail, a distal phalanx, and compliant pulp with force-transparent QDD actuation for force-regulated manipulation.
\item An energy-based fingertip model showing how material stiffness and contact curvature govern deformation partitioning between global bending and local contact indentation.
\item \rev{Ablation-based experimental studies showing how the distal phalanx, fingernail, and compliant pulp jointly shape contact behavior, contact stability, and  \rrev{fingernail-mediated} contact force transmission across diverse object geometries and contact conditions.}
\end{enumerate}

\section{Related Work}

\subsection{Dexterous Hand Design}
Robotic hand design has long balanced three competing demands: compact packaging, dexterous kinematics, and effective force interaction~\cite{Piazza2019AHands}. Many systems embed high-reduction actuators directly within the fingers~\cite{Shaw2023LEAPLearning,dmanus,tesolloDG5FHumanoid}, simplifying transmission and joint-state estimation but increasing distal mass, rigidity, and vulnerability to contact. The fixed housing geometry of such actuator modules also constrains finger morphology and limits overall hand form factor.

To mitigate distal inertia, other designs place actuation proximally and transmit motion through tendons~\cite{Kim2019FluidCapability,rollingcontacthand,orca,trx,pisa/iit,shadowrobotShadowDexterous}. These approaches support lightweight finger structures, but friction, hysteresis, and elastic elongation in the transmission reduce force-motion transparency and complicate state estimation, particularly in compact hands without distal joint encoders. Linkage-driven hands~\cite{Kim2021IntegratedHand,linkagehand,linkagereview,romero2024eyesight,Lin2022AManipulations} avoid tendon elasticity and can provide more direct mechanical coupling, but often introduce closed-chain kinematics, configuration-dependent transmission ratios, and packaging tradeoffs.

\rev{Within this design space, our work centers on a linkage-driven, proprioceptive hand architecture in which the finger mechanism and hybrid fingertip are co-designed for bidirectional transmission of motion and contact force. Rather than treating the fingertip as a passive contact element, we study how finger mechanics and fingertip substructure can be jointly designed to shape contact at the interface while preserving faithful transmission of contact-induced forces back to the proprioceptive actuation pathway.}

\subsection{Compliant Fingertips and Robotic Fingernails}
Compliant fingertips and soft contact interfaces have been widely explored to provide passive adaptation and robust interaction under geometric uncertainty. Existing approaches range from soft pads attached to rigid fingers to structured compliant fingertips and grasp-related contact interfaces that exploit distributed deformation during interaction~\cite{Crooks2016FinOptimization,Seo2025LEGATO:Tool,Chi2024UniversalRobots}. While these designs improve adaptability, fully compliant contacts can be disadvantageous for thin-object acquisition and precise contact initiation, where localized stiffness and well-defined contact geometry are beneficial.

\rev{To address this need, prior work has explored both tactile fingertip designs and robotic fingernails. Recent tactile fingertips have studied how distal interface geometry and sensing structure can support dexterous contact interaction~\cite{xu2025multimodaltactile,shang2025forte}}, while robotic fingernails have more directly targeted rigid distal-edge interaction for pinching and thin-object acquisition~\cite{Murakami2003NovelManipulation,Babin2018PickingGO}. Subsequent systems have used nail-like or structured distal features for low-clearance edge manipulation, tactile sensing, and object pickup from flat surfaces~\cite{Do2024DenseTact-Mini:Surfaces,Torrey2015ImprovingFingernails,jain2020nail,fang2025dexop}. In most of these works, however, the distal structure primarily serves as a grasp aid, sensing substrate, or task-specific interaction feature at the fingertip.


Our work builds on this literature but takes a different view of the fingertip structure. \rrev{While many prior designs approximate compliant \textit{None}-type or nail-assisted \textit{N-only}-type endpoints in Fig.~4(a), we study a hybrid fingertip in which the fingernail, embedded distal support, and compliant pulp jointly shape how contact is initiated, supported, and transmitted.}

\section{Design}
\rev{The PLATO Hand is designed around two coupled objectives: mechanically structuring fingertip contact and preserving force-motion transparency in the finger mechanism.} To this end, the hand integrates a hybrid fingertip composed of a rigid fingernail, a distal phalanx, and a compliant fingerpulp, together with a five-bar transmission driven by QDD actuators. This combination enables localized, stable contact at the fingertip and low-impedance, high-bandwidth force-regulated interaction at the hand level. The resulting platform comprises three fingers with eight fully actuated DoF for grasping and dexterous manipulation.

\begin{figure}[t]
  \centering
  \includegraphics[width=1.0\linewidth]{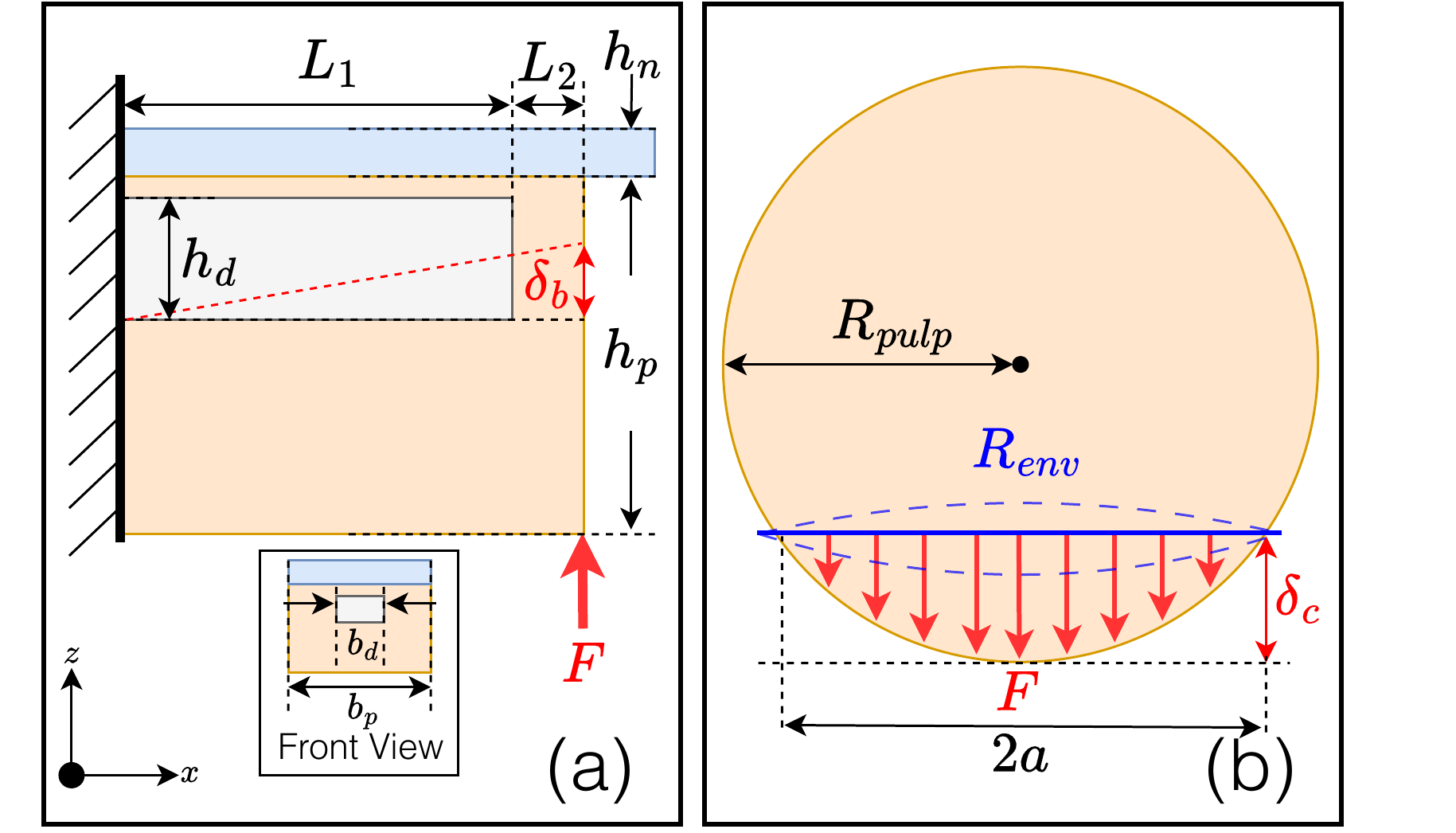}
  \caption{\textbf{Strain Energy Fingertip Model}.
 (a) Composite cantilever model of the PLATO Hand fingertip, consisting of a
  thin, rigid fingernail, soft pulp, and an embedded distal phalanx. The model
  captures bending deformation through a piecewise flexural rigidity
  determined by the layered geometry.
  (b) Hertzian contact model describing local indentation of the pulp against an external surface with radius $R_{\mathrm{env}}$. The indentation depth $\delta_c$ determines the contact radius. The total fingertip deformation couples beam bending and contact indentation: $\delta_{\mathrm{total}} = \delta_b + \delta_c$.
  }
  \label{fig:model}
\end{figure}


\subsection{Hybrid Fingertip Design}
\rev{The PLATO Hand employs a hybrid fingertip that combines a rigid fingernail, a distal phalanx, and a compliant fingerpulp to structure contact at the fingertip. This hybrid construction is designed to suppress global fingertip bending while preserving compliance at the contact interface.}

In this subsection, we analyze how the hybrid fingertip structure redistributes deformation from global bending to localized contact indentation.
We adopt an energy-based formulation that decomposes fingertip deformation into two components: bending of the fingertip structure and local indentation at the contact interface.
Under a prescribed displacement $\delta_{\text{total}}$, kinematic compatibility decomposes the deformation into bending and contact indentation, such that $\delta_{\text{total}} = \delta_b + \delta_c$. The total elastic energy then separates into
\begin{align}
    U_{\text{total}}(\delta_{\text{total}}) = U_b(\delta_b) + U_c(\delta_c).
\end{align}

The strain energy associated with each deformation mode is obtained from the work--energy principle by integrating the corresponding reaction force over its deformation.

\begin{figure*}[t]
  \centering
  \includegraphics[width=1.0\linewidth]{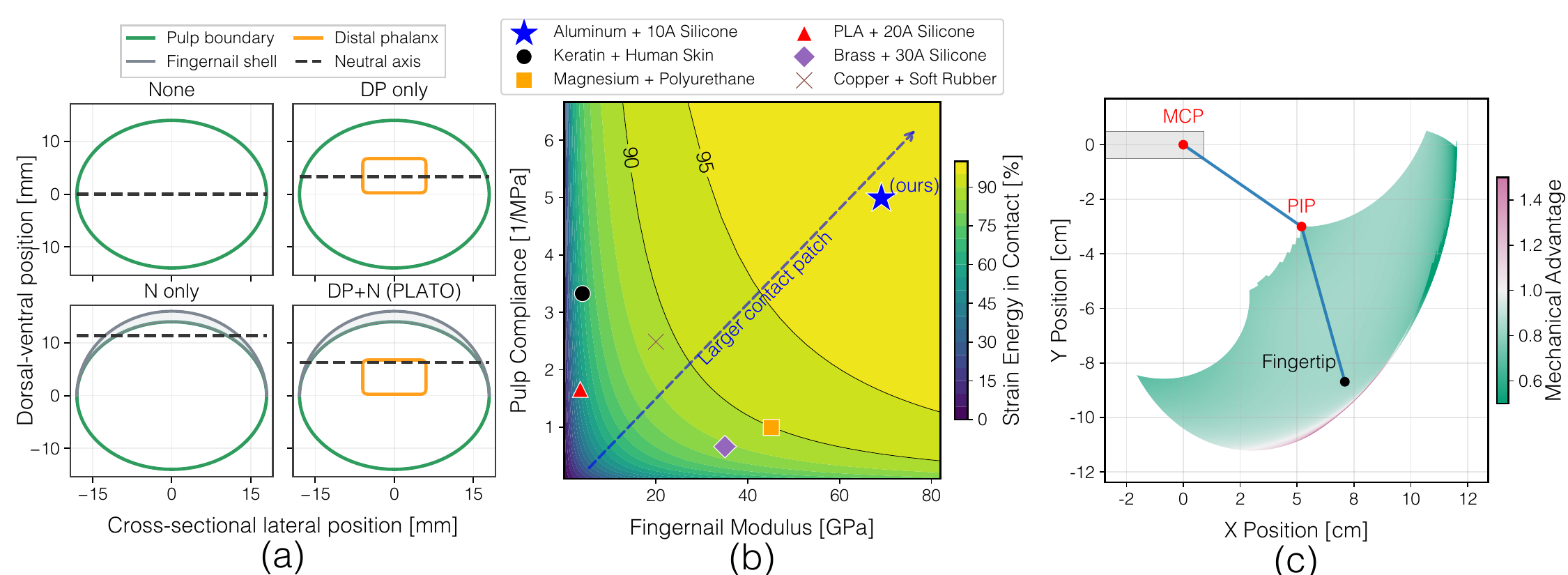}
  \caption{ 
     \textbf{Energy-based design characterization of the fingertip and finger kinematics.}
    (a) \rev{Cross-sectional comparison of fingertip design variants showing how the fingernail and distal phalanx modify the composite section. 
        The fingernail provides dorsal reinforcement, while the embedded distal phalanx provides internal support. Together, these features shift the neutral-axis location and increase the effective bending rigidity of the fingertip.}
    (b) Heatmap of the fraction of strain energy in contact indentation under a fixed fingertip approach. Higher nail stiffness and pulp compliance concentrate deformation at the contact rather than in global bending. The PLATO Hand design achieves over 95\% energy concentration in local indentation.
    (c) Fingertip Cartesian workspace and corresponding mechanical advantage distribution for the optimized linkage, showing reduced variation across the reachable workspace.
    }
  \label{fig:valid}
\end{figure*}

\subsubsection{Bending Energy}
We model the fingertip as a bonded layered beam of total length $L = L_1 + L_2$, where the segment spanning $0 \le x \le L_1$ includes the distal phalanx, pulp, and fingernail, and the segment $L_1 < x \le L$ consists of the pulp and fingernail alone. \rev{This partition reflects the composite architecture of the fingertip, with the distal phalanx present only over the proximal segment and the fingernail forming the stiff dorsal layer throughout.} The resulting flexural rigidity is piecewise constant:
\begin{align}
EI(x)=
\begin{cases}
(EI)_{1}, &
0 \le x \le L_1\quad
\text{\scriptsize (distal + pulp + nail)},\\[4pt]
(EI)_{2}, &
L_1 < x \le L \quad
\text{\scriptsize (pulp + nail)}.
\end{cases}
\end{align}
\rev{As illustrated in Fig.~\ref{fig:valid}(a), the fingernail and the distal phalanx modify the composite cross-section, shifting the neutral axis and increasing the effective flexural rigidity of the fingertip.}
To obtain a single equivalent stiffness, we define an effective bending rigidity $(EI)_{\mathrm{eff}}$ by equating the strain energy under a tip load to that of a uniform beam.
\begin{align}
U_b(\delta_b) 
= \int_{0}^{L} \frac{M(x)^2}{2 EI(x)}\,dx
= \frac{1}{2}\,\frac{3 (EI)_{\mathrm{eff}}}{L^3}\,\delta_b^2.
\end{align}

\subsubsection{Contact Indentation Energy}

We model the fingertip as a sphere contacting a surface under frictionless, non-adhesive, quasistatic loading. The soft pulp is treated as an elastic half-space with effective modulus $E^*=E_p/(1-\nu_p^2)$, where $E_p$ is the Young's modulus and $\nu_p$ is the Poisson ratio of the pulp material. 
The effective contact radius $R_{\mathrm{eff}}$ is determined by the combined curvatures of the fingertip and the environment, and reduces to the fingertip radius $R_{\mathrm{pulp}}$ for contact with a flat surface.

Under Hertzian contact, the force-indentation relationship and contact patch radius are given by
\begin{align}
F(\delta_c) = \frac{4}{3} E^* \sqrt{R_{\mathrm{eff}}}\,\delta_c^{3/2}, \quad
a = \sqrt{R_{\mathrm{eff}}\,\delta_c}.
\end{align}

Applying the work-energy principle, the contact strain energy is
\begin{align}
U_c(\delta_c)
= \int_0^{\delta_c} F\,d\delta
= \frac{8}{15} E^* \sqrt{R_{\mathrm{eff}}}\,\delta_c^{5/2}.
\end{align}

\subsubsection{Energy Minimization and Deformation Partitioning}

The internal deformation partition $(\delta_b,\delta_c)$ is determined by the fingertip structure through energy minimization. For a given externally imposed fingertip approach, the realized deformation is the one that minimizes the total elastic strain energy.

We model this as
\begin{align}
\min_{\delta_b,\delta_c} \quad
& U_b(\delta_b) + U_c(\delta_c) \label{eq:opt_a}\\
\text{subject to} \quad
& \delta_b + \delta_c = \delta_{\text{total}}. \label{eq:opt_b}
\end{align}
Solving the first-order optimality conditions gives
\begin{align}
\delta_c + \beta\,\delta_c^{3/2} = \delta_{\text{total}}, \quad
\beta = \frac{4 E^* \sqrt{R_{\mathrm{eff}}} L^3}{9 (EI)_{\mathrm{eff}}}.
\label{eqn:beta}
\end{align}

The parameter $\beta$ determines how the imposed approach is split between global bending and local contact indentation. Increasing $(EI)_{\mathrm{eff}}$ reduces $\beta$ and suppresses global bending, while the compliant pulp preserves local contact deformation. As a result, a larger portion of the imposed approach is directed into indentation at the contact interface.

Substituting the solution for $\delta_c$ into the energy expressions gives
\begin{align}
U_{\text{total}}
= \frac{1}{2}\,\frac{3 (EI)_{\mathrm{eff}}}{L^3}\,(\delta_{\text{total}} - \delta_c)^2
+ \frac{8}{15} E^* \sqrt{R_{\mathrm{eff}}}\,\delta_c^{5/2}.
\end{align}
The contact energy fraction is
\begin{align}
\eta_{\mathrm{contact}}
= \frac{U_c}{U_{\text{total}}}.
\end{align}
Increasing $(EI)_{\mathrm{eff}}$ shifts elastic energy from global bending into local contact deformation, increasing $\eta_{\mathrm{contact}}$ and enlarging $a$ for a given fingertip approach. \rev{In the proposed structure, this increase in effective bending rigidity arises from the combined contributions of the fingernail and the distal phalanx, which suppress global bending while allowing the surrounding soft pulp to remain the primary site of local contact deformation.}

\begin{figure*}
  \centering
  \includegraphics[width=\linewidth]{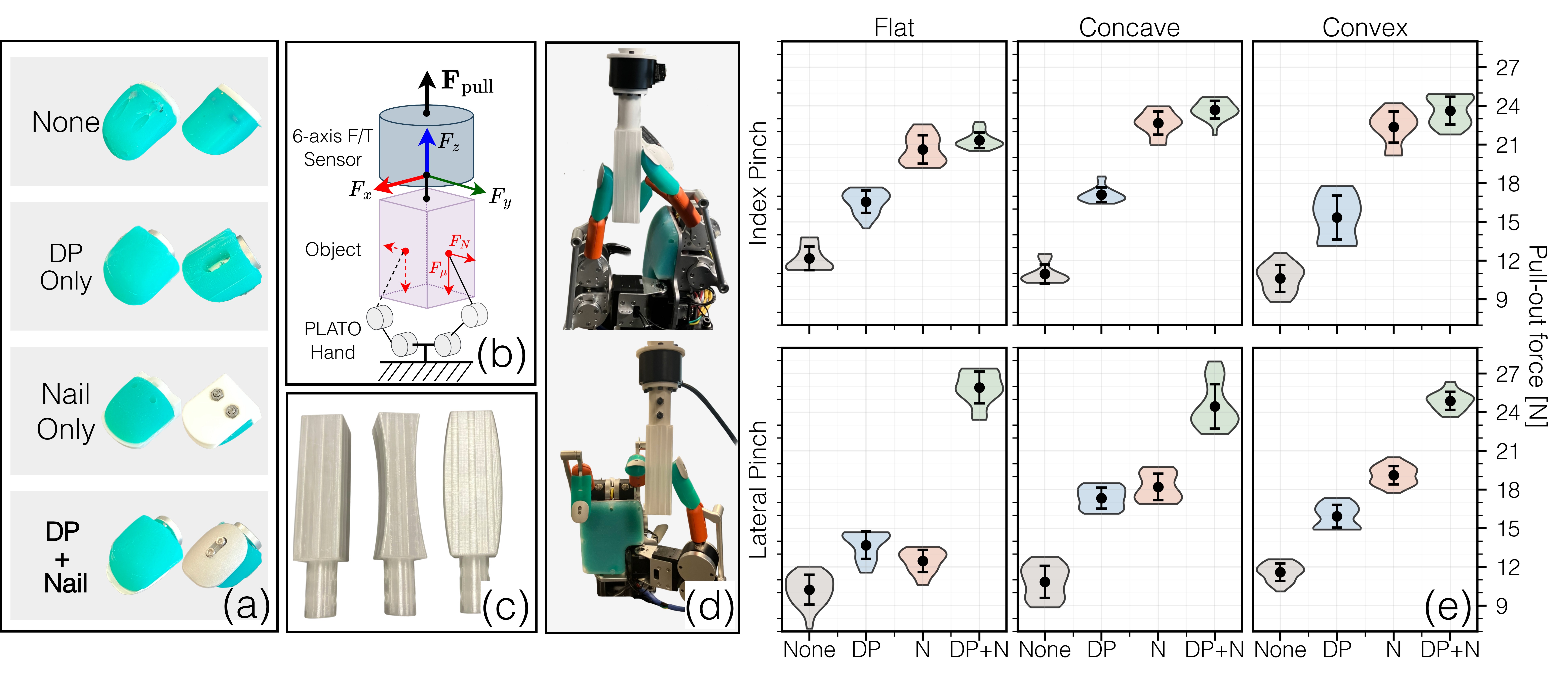}
  \caption{
\textbf{Hybrid fingertip architecture shapes contact mechanics across object geometries.}
  \rev{(a) Fingertip design variants used in the study: pulp-only (None), distal-phalanx-only (DP), fingernail-only (N), and the combined distal phalanx + fingernail design (DP+N). All fingertips are made from 10A silicone using injection molding.
  (b) Pull-out test setup used to evaluate pinch stability under tangential disturbance while measuring pull-out forces using a force-torque sensor.
  (c) 3D printed PETG test objects with flat, concave, and convex geometries.
  (d) Representative grasp configurations for index pinch and lateral pinch during the pull-out experiments.
  (e) Pull-out force across fingertip designs and object geometries for index pinch and lateral pinch (n=30). The results show that fingertip architecture systematically affects pinch stability, with the combined DP+N configuration yielding the strongest and most consistent performance across geometries.}
  }
  \label{fig:pullout}
\end{figure*}

\subsection{Finger Linkage Transmission Design}

The finger mechanism employs a five-bar linkage to transmit motion between the actuated joint and the distal joint without colocating actuators at the joints. Due to the linkage kinematics shown in Fig.~\ref{fig:design}(c), the relationship between the actuator angle and the distal joint angle is nonlinear and depends on the configuration variables $\phi_1$ and $\phi_2$. This kinematic nonlinearity results in a configuration-dependent mechanical advantage (MA) between the actuator angle $\phi_2$ and the joint angle $\phi_4$~\cite{Zi2011DynamicControl}.

Using loop-closure kinematics and Cramer's rule applied to the mechanism Jacobian, the differential relationship between $\phi_2$ and $\phi_4$ is
\begin{align}
N(\phi_2, \phi_4)
= \frac{d\phi_2}{d\phi_4}
= \frac{-L_4\sin(\phi_3 - \phi_4)}{L_2\sin(\phi_2 - \phi_3)},
\label{eq:ma_jacobian}
\end{align}
where $N(\phi_2, \phi_4)$ denotes the geometric mechanical advantage.

Large variations in MA introduce nonlinear distortions in the sensed force-torque relationship, degrading force control accuracy. We therefore optimize the link lengths $L_1$ and $L_3$ to reduce MA variation across the configuration space while satisfying geometric and workspace constraints.

The optimization problem is formulated as
\begin{subequations}
\begin{align}
\min_{L_1, L_3} \quad
    & J = w_1\|\Delta\boldsymbol{\theta}\|_2
        + w_2\sigma_N
        + w_3\mathrm{TI}
        + w_4\|\mathbf{L}\|_2^2
        \label{eq:opt_obj}
\\
\text{s.t.}\quad
    & \sum_{i=1}^{5} L_i e^{j\phi_i} = 0
    \label{eq:opt_loop}
\\
    & \mathbf{L}_{\min} \le \mathbf{L} \le \mathbf{L}_{\max},
    \label{eq:opt_bounds}
\end{align}
\end{subequations}
where the objective terms are defined as follows:
\begin{itemize}
    \item $\|\Delta\boldsymbol{\theta}\|_2$: reachable joint-angle workspace coverage.
    \item $\sigma_N$: standard deviation of $N(\phi_2,\phi_4)$ over the sampled workspace.
    \item $\mathrm{TI}$: transmission index penalizing near-singular linkage configurations.
    \item $\|\mathbf{L}\|_2^2$: link length regularization.
\end{itemize}
The constraints enforce geometric closure of the linkage and restrict link lengths to physically feasible ranges.

The resulting nonlinear optimization problem is solved using the Covariance Matrix Adaptation Evolution Strategy (CMA-ES), which is suitable for non-convex design spaces with multiple local minima~\cite{hansen2023cmaevolutionstrategytutorial}. The resulting workspace and MA distribution are shown in Fig.~\ref{fig:valid}(c).

\subsection{Hand Topology Design}
The PLATO Hand is formed by assembling three instances of the finger design described above, each actuated by 1:8 QDD actuators.
The thumb includes two additional proximal degrees of freedom actuated by compact servo motors, while the finger mechanisms drive the remaining joints.
This topology supports a range of common three-finger grasping and manipulation configurations~\cite{Cutkosky1989OnTasks}.

\begin{figure*}
  \centering
  \includegraphics[width=\linewidth]{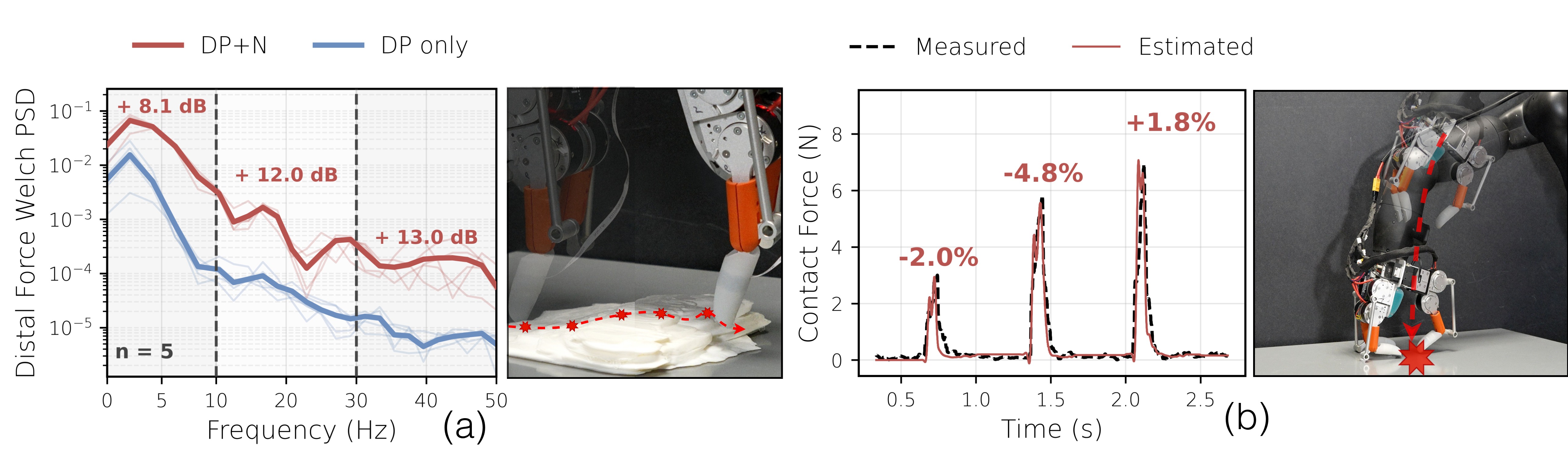}
  \caption{
  \textbf{\rrev{Dorsal-contact force transmission and proprioceptive estimation.}}
    (a) Power spectral density (PSD) of the contact force measured by the fingertip force/torque sensor while the dorsal nail side of the fingertip traces a textured surface (n=5). The DP+N fingertip exhibits higher spectral power than the DP-only fingertip across the measured frequency range, indicating improved transmission of dorsal-contact-induced force variations. The annotated dB values denote the increase in band-integrated PSD power for each frequency range.
    \rrev{(b) Measured dorsal-contact} force magnitude and proprioceptive force estimate during repeated impact interactions between the dorsal nail side of the fingertip and a rigid surface (n=3). The annotated percentages denote signed peak error for the three impact events.
    }

  \label{fig:sense}
\end{figure*}

\section{Experiments}
We evaluate how the mechanical design of the PLATO Hand shapes contact behavior during manipulation. Using the PLATO Hand mounted on a Roboligent Optimo 7-DoF robotic manipulator,  we investigate three aspects: (1) pinch stability across object geometries, (2) transmission and estimation of contact-induced force variations, and (3) performance in contact-rich manipulation tasks.

\subsection{Pull-Out Evaluation Across Fingertip Architectures}
\rev{
To evaluate pinch stability under tangential disturbance, we performed pull-out experiments across four fingertip architectures: pulp-only (None), distal-phalanx-only (DP), fingernail-only (N), and the combined distal phalanx plus fingernail design (DP+N). As shown in Fig.~\ref{fig:pullout}(a), these variants isolate the structural contributions of the compliant pulp, distal phalanx, and fingernail. The evaluation was conducted on flat, concave, and convex objects (Fig.~\ref{fig:pullout}(c)) using index-pinch and lateral-pinch configurations (Fig.~\ref{fig:pullout}(d)). For each condition, the PLATO Hand applied a comparable normal preload under joint impedance control, and the object was manually pulled until grasp failure, while the maximum pull-out force was recorded using a Rokubi 6-axis force--torque sensor. The test setup is shown in Fig.~\ref{fig:pullout}(b).
}

\rev{
The results in Fig.~\ref{fig:pullout}(e) show that fingertip architecture affects pull-out stability \rrev{in a grasp-dependent manner}. In index pinch, the N and DP+N variants achieve similarly high pull-out forces across all tested surface geometries, indicating that distal-edge reinforcement is already a dominant factor in this grasp mode. In contrast, the \rrev{combined DP+N structure provides its clearest advantage in lateral pinch, where it} consistently outperforms the other architectures across all object geometries.
}

\rev{
Relative to the N-only configuration, \rrev{DP+N increases lateral-pinch pull-out force by approximately} 108\% on the flat object, 34\% on the concave object, and 30\% on the convex object. Relative to the pulp-only condition, the \rrev{corresponding improvements are approximately} 153\%, 126\%, and 116\% for the flat, concave, and convex objects, respectively. 
}

\rev{
This trend is consistent with the design analysis in Fig.~\ref{fig:valid}. The fingernail and distal phalanx jointly shift the neutral axis and increase the effective bending rigidity of the fingertip, thereby suppressing global bending while preserving local compliance at the pulp. In index pinch, where distal-edge interaction already dominates the contact, the fingernail alone captures much of this benefit. In lateral pinch, however, stable pull-out requires not only edge engagement but also mechanically supported contact under off-axis tangential loading, making the combined DP+N structure especially advantageous. 
}

\rrev{\subsection{Dorsal-Contact Force Sensing Characterization}}
\rrev{We evaluate how the hybrid fingertip affects fingernail-mediated force transmission and proprioceptive force estimation under dorsal contact.}
\rev{In both experiments, the robotic arm executed a replayed reference motion recorded through kinesthetic teaching, while the robotic hand was controlled with fixed joint impedance. Distal force--torque measurements were used as the reference sensing modality, and force estimates were derived from motor current using the linkage transmission model and the finger Jacobian.
}

\subsubsection{Contact Force Sensitivity}
To characterize contact force transmission, the fingertip was swept across a textured surface. 
\rev{Distal force measurements were recorded during this motion, and spectral power was compared between the DP-only and DP+N fingertip configurations across three frequency bands.
As shown in Fig.~\ref{fig:sense}(a), the DP+N fingertip exhibits consistently higher spectral energy across the measured frequency range. This indicates that the rigid fingernail more effectively transmits contact-induced force variations, whereas the fingertip without the nail attenuates them through distributed deformation and material damping.
}

\subsubsection{Proprioceptive Force Estimation}
To evaluate \rrev{proprioceptive force estimation during dynamic nail-mediated contact, we compared motor-current-based force estimates against distal force measurements} during three repeated impact interactions between \rrev{the dorsal nail side of the fingertip} and a rigid surface. For the traces shown in Fig.~\ref{fig:sense}(b), a scalar Kalman filter was applied for smoothing.

As shown in Fig.~\ref{fig:sense}(b), the proprioceptive force estimates track the measured contact force magnitude closely across repeated impact events, with small signed peak errors in all three impacts.
This suggests that the low mechanical impedance of the finger mechanism supports reliable force estimation during \rrev{rapid dorsal contact mediated by the fingernail}.


\begin{figure*}[t]
  \centering
  \includegraphics[width=1\linewidth]{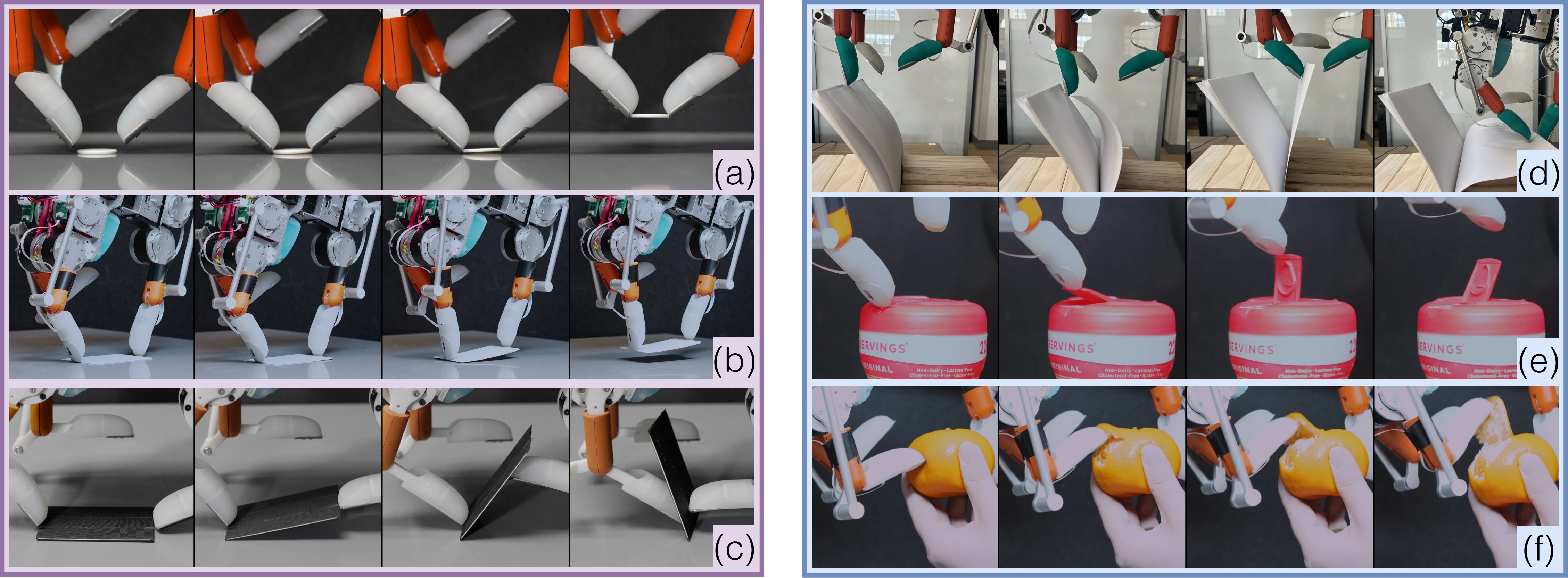}
  \caption{\textbf{Teleoperated manipulation tasks enabled by the PLATO Hand.}
    (a) Coin picking and (b) card picking --- retrieving thin objects from a flat surface;
    (c) card flipping --- initiating rotation and grasping a card;
    (d) paper singulation --- isolating and pulling a single sheet from a stack;
    (e) lid opening --- prying open a sealed container;
    (f) orange peeling --- puncturing and following the peel along a curved fruit surface.
    These tasks illustrate how mechanically structured contact and proprioceptive actuation support edge-sensitive and contact-rich manipulation across diverse object geometries.}
  \label{fig:tasks}
\end{figure*}

\subsection{Manipulation Task Demonstrations}
We evaluate the PLATO Hand on manipulation tasks that require structured contact geometry together with force-aware interaction. 
\rev{The robotic arm is teleoperated using a SpaceMouse under Cartesian impedance control, while the hand joints are controlled with fixed impedance. Task progression is governed by a contact-dependent state machine that triggers task-specific, predefined hand joint trajectories. For edge-interaction tasks, the operator first teleoperates the arm to establish the desired contact configuration, after which the corresponding predefined hand joint trajectory is triggered. For contact-triggered tasks such as coin and card picking, proprioceptive force sensing is used to detect contact and trigger the corresponding hand joint trajectory.}

\begin{table}[t]
\centering
\vspace{4pt}
\caption{\textbf{Manipulation Tasks Success Rates}}
\label{tab:plato_hand_nail_ablation}
\resizebox{0.95\columnwidth}{!}{%
\begin{tabular}{lcc}
\toprule
\textbf{Task} & \textbf{Distal Phalanx only} & \textbf{Distal Phalanx + Fingernail} \\
\midrule
\quad Paper singulation   & \;\,0/10 & \textbf{8/10} \\
\quad Lid opening         & \;\,4/10 & \textbf{9/10} \\
\quad Orange peeling      & \;\,0/3  & \textbf{3/3} \\
\midrule
\quad Coin picking        & \;\,0/10 & \textbf{9/10} \\
\quad Card picking        & \;\,0/10 & \textbf{10/10} \\
\quad Card flipping       & \;\,0/10 & \textbf{8/10} \\
\bottomrule
\end{tabular}
}
\end{table}

Table~\ref{tab:plato_hand_nail_ablation} reports success rates with DP-only and DP+N fingertips. 
\rev{A trial was considered unsuccessful if the task was not completed within 1~min; for coin and card tasks, dropping the object off the table was also counted as a failure. These demonstrations indicate that the fingernail primarily improves the repeatability of contact initiation and edge stabilization.}

\subsubsection{Rigid Edge Interaction Tasks}
These tasks require the fingertip to establish a rigid edge contact and transmit localized forces through a small contact area:
\begin{itemize}
    \item \textbf{Paper singulation}: wedge interaction with sub-millimeter gaps between stacked A4 papers.
    \item \textbf{Lid opening}: edge contact and torque transmission at a predefined leverage point.
    \item \textbf{Orange peeling}: concentrated force application to puncture and tear the surface.
\end{itemize}
Without the fingernail, these tasks often fail because the compliant pulp deforms before stable edge contact is established. \rev{This blunts the intended wedge or leverage geometry, especially in paper singulation and orange peeling. Lid opening is less restrictive, so occasional success is still possible when the initial edge contact is favorable.} With the fingernail, the required contact geometry is maintained, leading to substantially higher success rates.

\subsubsection{Contact-Rich Manipulation Tasks}
These tasks involve repeated contact transitions under low-force interaction:
\begin{itemize}
    \item \textbf{Coin picking}: lift initiation on a rigid, low-profile object.
    \item \textbf{Card picking}: edge interaction and stabilization of a thin, flexible object.
    \item \textbf{Card flipping}: torque application at an object edge during controlled rotation.
\end{itemize}
Without the fingernail, these tasks consistently fail because pulp deformation disrupts contact initiation and edge stabilization. \rev{In particular, the soft pulp broadens the initial contact and makes it difficult to lift, trap, or pivot the object reliably.} With the fingernail, the fingertip maintains a more stable contact geometry, enabling successful execution. These results show that manipulation capability in the PLATO Hand emerges from the integration of mechanically structured contact and proprioceptive, force-regulated actuation.



\section{Conclusion}
In this work, we introduced the PLATO Hand, a proprioceptive robotic hand that demonstrates how fingertip structure can be designed to \rrev{create stable and task-relevant contact conditions} for manipulation. By integrating a rigid fingernail, embedded distal support, and compliant pulp, the proposed hybrid fingertip \rrev{mechanically organizes how contact is initiated, supported, and transmitted at the fingertip}. \rrev{Rather than acting merely as a soft contact surface, the fingertip functions as a structured contact interface that promotes stable edge interaction, preserves local compliance where needed, and improves fingernail-mediated dorsal-contact force transmission.}

\rev{The proposed bending--indentation model, based on Hertzian contact, is useful for explaining how deformation energy is partitioned, but remains limited under large deformation, friction, and other nonlinear contact effects.}
\rrev{The current linkage-driven mechanism also highlights a compactness challenge: its exposed transmission layout can be susceptible to unintended environmental contact.}
\rev{Future work will extend this framework toward richer contact models and more compact, contact-aware linkage designs for higher-DoF hands.}


\section*{Acknowledgment}
This work was supported by Sony Group Corporation.
We sincerely thank Roboligent for its generous hardware support.

\newpage
\renewcommand{\baselinestretch}{1.0}
\footnotesize
\bibliographystyle{IEEEtran}
\bibliography{references}

@inproceedings{utahmit,
  author    = {Jacobsen, Stephen C. and Iversen, Edwin K. and Knutti, David F. and Johnson, R. Todd and Biggers, Klaus B.},
  title     = {Design of the Utah/{M.I.T.} Dextrous Hand},
  booktitle = {Proc. IEEE Int. Conf. Robot. Autom. (ICRA)},
  year      = {1986},
  volume    = {3},
  pages     = {1520--1532}
}

@inproceedings{robonaut,
  author    = {Lovchik, Chris S. and Diftler, Myron A.},
  title     = {Robonaut Hand: A Dexterous Robot Hand for Space},
  booktitle = {Proc. IEEE Int. Conf. Robot. Autom. (ICRA)},
  year      = {1999},
  volume    = {2},
  pages     = {907--912},
  doi       = {10.1109/ROBOT.1999.772420}
}

@article{gifuhand,
  author  = {Kawasaki, H. and Komatsu, T. and Uchiyama, K.},
  title   = {Dexterous Anthropomorphic Robot Hand with Distributed Tactile Sensor: {Gifu} Hand {II}},
  journal = {IEEE/ASME Trans. Mechatronics},
  year    = {2002},
  volume  = {7},
  number  = {3},
  pages   = {296--303},
  doi     = {10.1109/TMECH.2002.802720}
}

@inproceedings{dlrhand,
  author    = {Liu, Hong and Wu, Kai and Meusel, Peter and Seitz, Norbert and Hirzinger, Gerd and Jin, Meng H. and Liu, Yu W. and Fan, Shao W. and Lan, Tian and Chen, Zhi P.},
  title     = {Multisensory Five-Finger Dexterous Hand: The {DLR}/{HIT} Hand {II}},
  booktitle = {Proc. IEEE/RSJ Int. Conf. Intell. Robots Syst. (IROS)},
  year      = {2008},
  pages     = {3692--3697},
  doi       = {10.1109/IROS.2008.4650624}
}

@article{acthand,
  author  = {Deshpande, Ashish D. and Xu, Zhe and Van der Weghe, Michael J. and Brown, Benjamin H. and Ko, Jonathan and Chang, Lillian Y. and Wilkinson, David D. and Bidic, Sean M. and Matsuoka, Yoky},
  title   = {Mechanisms of the Anatomically Correct Testbed Hand},
  journal = {IEEE/ASME Trans. Mechatronics},
  year    = {2013},
  volume  = {18},
  number  = {1},
  pages   = {238--250},
  doi     = {10.1109/TMECH.2011.2166801}
}

@article{Kim2019FluidCapability,
  author  = {Kim, Yong-Jae and Yoon, Junsuk and Sim, Young-Woo},
  title   = {Fluid Lubricated Dexterous Finger Mechanism for Human-Like Impact Absorbing Capability},
  journal = {IEEE Robot. Autom. Lett.},
  year    = {2019},
  volume  = {4},
  number  = {4},
  pages   = {3971--3978}
}

@inproceedings{rollingcontacthand,
  author    = {Toshimitsu, Yasunori and Forrai, Benedek and Cangan, Barnabas Gavin and Steger, Ulrich and Knecht, Manuel and Weirich, Stefan and Katzschmann, Robert K.},
  title     = {Getting the Ball Rolling: Learning a Dexterous Policy for a Biomimetic Tendon-Driven Hand with Rolling Contact Joints},
  booktitle = {Proc. IEEE-RAS Int. Conf. Humanoid Robots (Humanoids)},
  year      = {2023},
  pages     = {1--7},
  doi       = {10.1109/Humanoids57100.2023.10375231}
}

@misc{orca,
  author        = {Christoph, Clemens C. and Eberlein, Maximilian and Katsimalis, Filippos and Roberti, Arturo and Sympetheros, Aristotelis and Vogt, Michel R. and Liconti, Davide and Yang, Chenyu and Cangan, Barnabas Gavin and Hinchet, Ronan J. and Katzschmann, Robert K.},
  title         = {{ORCA}: An Open-Source, Reliable, Cost-Effective, Anthropomorphic Robotic Hand for Uninterrupted Dexterous Task Learning},
  year          = {2025},
  note          = {arXiv:2504.04259}
}

@inproceedings{trx,
  author    = {Yang, Sicheng and Lee, Wang Wei and Zhang, Zhong and Xiong, Youda and Liang, Jiaming and Lu, Peng and Zhu, Yonghui and Liu, Tianliang and Li, Jingchen and Wang, Rui and Li, Xiong and Zheng, Yu},
  title     = {{TRX-Hand5}: An Anthropomorphic Hand with Integrated Tactile Feedback for Grasping and Manipulation in Human Environments},
  booktitle = {Proc. IEEE/RSJ Int. Conf. Intell. Robots Syst. (IROS)},
  year      = {2024},
  pages     = {5289--5296},
  doi       = {10.1109/IROS58592.2024.10801666}
}

@article{pisa/iit,
  author  = {Santina, Cosimo Della and Piazza, Cristina and Grioli, Giorgio and Catalano, Manuel G. and Bicchi, Antonio},
  title   = {Toward Dexterous Manipulation with Augmented Adaptive Synergies: The Pisa/{IIT} {SoftHand} 2},
  journal = {IEEE Trans. Robot.},
  year    = {2018},
  volume  = {34},
  number  = {5},
  pages   = {1141--1156},
  doi     = {10.1109/TRO.2018.2830407}
}

@misc{shadowrobotShadowDexterous,
  author       = {{Shadow Robot Company}},
  title        = {Shadow Dexterous Hand Series -- Research and Development Tool},
  howpublished = {\url{https://shadowrobot.com/dexterous-hand-series/}}
}

@article{Kim2021IntegratedHand,
  author  = {Kim, Uikyum and Jung, Dawoon and Jeong, Heeyoen and Park, Jongwoo and Jung, Hyun Mok and Cheong, Joono and Choi, Hyouk Ryeol and Do, Hyunmin and Park, Chanhun},
  title   = {Integrated Linkage-Driven Dexterous Anthropomorphic Robotic Hand},
  journal = {Nature Commun.},
  year    = {2021},
  volume  = {12},
  number  = {1}
}

@ARTICLE{linkagehand,
  author  = {Li, Guotao and Liang, Xu and Gao, Yifan and Su, Tingting and Liu, Zhijie and Hou, Zeng-Guang},
  title   = {A Linkage-Driven Underactuated Robotic Hand for Adaptive Grasping and In-Hand Manipulation},
  journal = {IEEE Trans. Autom. Sci. Eng.},
  year    = {2024},
  volume  = {21},
  number  = {3},
  pages   = {3039--3051},
  doi     = {10.1109/TASE.2023.3273721}
}

@article{dmanus,
  author  = {Bhirangi, Raunaq and DeFranco, Abigail and Adkins, Jacob and Majidi, Carmel and Gupta, Abhinav and Hellebrekers, Tess and Kumar, Vikash},
  title   = {All the Feels: A Dexterous Hand with Large-Area Tactile Sensing},
  journal = {IEEE Robot. Autom. Lett.},
  year    = {2023},
  volume  = {8},
  number  = {12},
  pages   = {8311--8318},
  doi     = {10.1109/LRA.2023.3327619}
}

@misc{allegrohand,
  author       = {{Wonik Robotics}},
  title        = {Allegro Hand},
  howpublished = {\url{https://www.allegrohand.com/}}
}

@misc{tesolloDG5FHumanoid,
  author       = {{Tesollo}},
  title        = {{DG}-5{F} Humanoid Robotic Hand for Dexterous Manipulation},
  howpublished = {\url{https://en.tesollo.com/dg-5f/}}
}

@article{Wensing2017ProprioceptiveRobots,
  author  = {Wensing, Patrick M. and Wang, Albert and Seok, Sangok and Otten, David and Lang, Jeffrey and Kim, Sangbae},
  title   = {Proprioceptive Actuator Design in the {MIT} Cheetah: Impact Mitigation and High-Bandwidth Physical Interaction for Dynamic Legged Robots},
  journal = {IEEE Trans. Robot.},
  year    = {2017},
  volume  = {33},
  number  = {3},
  pages   = {509--522}
}

@inproceedings{Sim2021TheRobots,
  author    = {Sim, Youngwoo and Ramos, Joao},
  title     = {The Dynamic Effect of Mechanical Losses of Transmissions on the Equation of Motion of Legged Robots},
  booktitle = {Proc. IEEE Int. Conf. Robot. Autom. (ICRA)},
  year      = {2021},
  pages     = {1191--1197}
}

@inproceedings{Jeong2024BaRiFlex:Learning,
  author    = {Jeong, Gu-Cheol and Bahety, Arpit and Pedraza, Gabriel and Deshpande, Ashish D. and Mart{\'i}n-Mart{\'i}n, Roberto},
  title     = {{BaRiFlex}: A Robotic Gripper with Versatility and Collision Robustness for Robot Learning},
  booktitle = {Proc. IEEE/RSJ Int. Conf. Intell. Robots Syst. (IROS)},
  year      = {2024},
  pages     = {4106--4113}
}

@inproceedings{romero2024eyesight,
  author    = {Romero, Branden and Fang, Hao-Shu and Agrawal, Pulkit and Adelson, Edward},
  title     = {Eyesight Hand: Design of a Fully-Actuated Dexterous Robot Hand with Integrated Vision-Based Tactile Sensors and Compliant Actuation},
  booktitle = {Proc. IEEE/RSJ Int. Conf. Intell. Robots Syst. (IROS)},
  year      = {2024}
}

@inproceedings{Lin2022AManipulations,
  author    = {Lin, Shiyuan and Zhao, Yuntian and Zhu, Zheng and Jia, Zhenzhong},
  title     = {A Quasi-Direct Drive Robot Hand for Reactive and Contact-Rich Manipulations},
  booktitle = {Proc. IEEE Int. Conf. Adv. Robot. Mechatronics (ICARM)},
  year      = {2022},
  pages     = {180--186},
  doi       = {10.1109/ICARM54641.2022.9959623}
}

@inproceedings{okamuradext,
  author    = {Okamura, A. M. and Smaby, N. and Cutkosky, M. R.},
  title     = {An Overview of Dexterous Manipulation},
  booktitle = {Proc. IEEE Int. Conf. Robot. Autom. (ICRA)},
  year      = {2000},
  volume    = {1},
  pages     = {255--262},
  doi       = {10.1109/ROBOT.2000.844067}
}

@article{rdexmani,
  author  = {Bullock, Ian M. and Ma, Raymond R. and Dollar, Aaron M.},
  title   = {A Hand-Centric Classification of Human and Robot Dexterous Manipulation},
  journal = {IEEE Trans. Haptics},
  year    = {2013},
  volume  = {6},
  number  = {2},
  pages   = {129--144},
  doi     = {10.1109/TOH.2012.53}
}

@inproceedings{Piraccini2014NailClinician,
  author    = {Piraccini, Bianca Maria},
  title     = {Nail Anatomy and Physiology for the Clinician},
  booktitle = {Nail Disorders},
  publisher = {Springer},
  year      = {2014},
  pages     = {1--6}
}

@inproceedings{Kumagai2022ComparisonNails,
  author    = {Kumagai, Ayane and Obata, Yoshinobu and Yabuki, Yoshiko and Jiang, Yinlai and Yokoi, Hiroshi and Togo, Shunta},
  title     = {Comparison of Precision Grasping Performance between Artificial Fingers with and without Nails},
  booktitle = {Proc. IEEE Global Conf. Life Sci. Technol. (LifeTech)},
  year      = {2022},
  pages     = {380--381}
}

@article{Shirato2017EffectDexterity,
  author  = {Shirato, Rikiya and Abe, Atsumi and Tsuchiya, Hikaru and Honda, Mizuki},
  title   = {Effect of Fingernail Length on the Hand Dexterity},
  journal = {J. Phys. Ther. Sci.},
  year    = {2017},
  volume  = {29},
  number  = {11},
  pages   = {1914--1919}
}

@misc{fang2025dexop,
  author        = {Fang, Hao-Shu and Romero, Branden and Xie, Yichen and Hu, Arthur and Huang, Bo-Ruei and Alvarez, Juan and Kim, Matthew and Margolis, Gabriel and Anbarasu, Kavya and Tomizuka, Masayoshi and Adelson, Edward and Agrawal, Pulkit},
  title         = {{DEXOP}: A Device for Robotic Transfer of Dexterous Human Manipulation},
  year          = {2025},
  note          = {arXiv:2509.04441}
}

@inproceedings{Do2024DenseTact-Mini:Surfaces,
  author    = {Do, Won Kyung and Dhawan, Ankush Kundan and Kitzmann, Mathilda and Kennedy, Monroe},
  title     = {{DenseTact-Mini}: An Optical Tactile Sensor for Grasping Multi-Scale Objects from Flat Surfaces},
  booktitle = {Proc. IEEE Int. Conf. Robot. Autom. (ICRA)},
  year      = {2024},
  pages     = {6928--6934}
}

@inproceedings{jain2020nail,
  author    = {Jain, S. and Stalin, T. and Subramaniam, V. and Agarwal, J. and Alvarado, P. Valdivia Y.},
  title     = {A Soft Gripper with Retractable Nails for Advanced Grasping and Manipulation},
  booktitle = {Proc. IEEE Int. Conf. Robot. Autom. (ICRA)},
  year      = {2020},
  pages     = {6928--6934},
  doi       = {10.1109/ICRA40945.2020.9197259}
}

@article{Odhner2013ACU,
  author  = {Odhner, Lael and Jentoft, Leif P. and Claffee, Mark R. and Corson, Nick and Tenzer, Yaroslav and Ma, Raymond R. and Buehler, Martin and Kohout, Robert C. and Howe, Robert D. and Dollar, Aaron M.},
  title   = {A Compliant, Underactuated Hand for Robust Manipulation},
  journal = {Int. J. Robot. Res.},
  year    = {2013},
  volume  = {33},
  pages   = {736--752}
}

@inproceedings{Murakami2003NovelManipulation,
  author    = {Murakami, K. and Hasegawa, T.},
  title     = {Novel Fingertip Equipped with Soft Skin and Hard Nail for Dexterous Multi-Fingered Robotic Manipulation},
  booktitle = {Proc. IEEE Int. Conf. Robot. Autom. (ICRA)},
  year      = {2003},
  volume    = {1},
  pages     = {708--713}
}

@article{Piazza2019AHands,
  author  = {Piazza, C. and Grioli, G. and Catalano, M. G. and Bicchi, A.},
  title   = {A Century of Robotic Hands},
  journal = {Annu. Rev. Control Robot. Auton. Syst.},
  year    = {2019},
  volume  = {2},
  pages   = {1--32},
  doi     = {10.1146/annurev-control-060117-105003}
}

@inproceedings{Shaw2023LEAPLearning,
  author    = {Shaw, Kenneth and Agarwal, Ananye and Pathak, Deepak},
  title     = {{LEAP} Hand: Low-Cost, Efficient, and Anthropomorphic Hand for Robot Learning},
  booktitle = {Robotics: Science and Systems (RSS)},
  year      = {2023}
}

@article{linkagereview,
  author  = {Kashef, S. Reza and Amini, Samane and Akbarzadeh, Alireza},
  title   = {Robotic Hand: A Review on Linkage-Driven Finger Mechanisms of Prosthetic Hands and Evaluation of the Performance Criteria},
  journal = {Mechanism Mach. Theory},
  year    = {2020},
  volume  = {145},
  pages   = {103677},
  doi     = {10.1016/j.mechmachtheory.2019.103677}
}

@article{Crooks2016FinOptimization,
  author  = {Crooks, Whitney and Vukasin, Gabrielle and O'Sullivan, Maeve and Messner, William and Rogers, Chris},
  title   = {{Fin Ray\textregistered} Effect Inspired Soft Robotic Gripper: From the {RoboSoft} Grand Challenge toward Optimization},
  journal = {Front. Robot. AI},
  year    = {2016},
  volume  = {3},
  pages   = {220991}
}

@article{Seo2025LEGATO:Tool,
  author  = {Seo, Mingyo and Park, H. Andy and Yuan, Shenli and Zhu, Yuke and Sentis, Luis},
  title   = {{LEGATO}: Cross-Embodiment Imitation Using a Grasping Tool},
  journal = {IEEE Robot. Autom. Lett.},
  year    = {2025},
  pages   = {1--8},
  doi     = {10.1109/LRA.2025.3535182}
}

@inproceedings{Chi2024UniversalRobots,
  author    = {Chi, Cheng and Xu, Zhenjia and Pan, Chuer and Cousineau, Eric and Burchfiel, Benjamin and Feng, Siyuan and Tedrake, Russ and Song, Shuran},
  title     = {Universal Manipulation Interface: In-The-Wild Robot Teaching without In-The-Wild Robots},
  booktitle = {Robotics: Science and Systems (RSS)},
  year      = {2024}
}

@misc{xu2025multimodaltactile,
  author = {Xu, Zhuowei and Si, Zilin and Zhang, Kevin and Kroemer, Oliver and Temel, Zeynep},
  title  = {A Multi-Modal Tactile Fingertip Design for Robotic Hands to Enhance Dexterous Manipulation},
  year   = {2025},
  note   = {arXiv:2510.05382}
}

@misc{shang2025forte,
  author = {Shang, Siqi and Seo, Mingyo and Zhu, Yuke and Chin, Lillian},
  title  = {{FORTE}: Tactile Force and Slip Sensing on Compliant Fingers for Delicate Manipulation},
  year   = {2025},
  note   = {arXiv:2506.18960}
}

@article{Babin2018PickingGO,
  author  = {Babin, Vincent and Gosselin, Cl{\'e}ment},
  title   = {Picking, Grasping, or Scooping Small Objects Lying on Flat Surfaces: A Design Approach},
  journal = {Int. J. Robot. Res.},
  year    = {2018},
  volume  = {37},
  pages   = {1484--1499}
}

@techreport{Torrey2015ImprovingFingernails,
  author = {Torrey, J. S. and Morrow, J. F. and Larkins, R. D. and Dang, S. T.},
  title  = {Improving Soft Pneumatic Actuator Fingers through Integration of Soft Sensors, Position and Force Control, and Rigid Fingernails},
  year   = {2015}
}

@article{Zi2011DynamicControl,
  author  = {Zi, Bin and Cao, Jianbin and Zhu, Zhencai},
  title   = {Dynamic Simulation of Hybrid-Driven Planar Five-Bar Parallel Mechanism Based on {SimMechanics} and Tracking Control},
  journal = {Int. J. Adv. Robot. Syst.},
  year    = {2011},
  volume  = {8},
  number  = {4},
  pages   = {28--33}
}

@misc{hansen2023cmaevolutionstrategytutorial,
  author = {Hansen, Nikolaus},
  title  = {The {CMA} Evolution Strategy: A Tutorial},
  year   = {2023},
  note   = {arXiv:1604.00772}
}

@article{Cutkosky1989OnTasks,
  author  = {Cutkosky, Mark R.},
  title   = {On Grasp Choice, Grasp Models, and the Design of Hands for Manufacturing Tasks},
  journal = {IEEE Trans. Robot. Autom.},
  year    = {1989},
  volume  = {5},
  number  = {3},
  pages   = {269--279}
}


\end{document}